\documentclass{article}

\usepackage{arxiv}
\usepackage{graphicx}%
\usepackage{multirow}%
\usepackage{amsmath,amssymb,amsfonts}%
\usepackage{amsthm}%
\usepackage{mathrsfs}%
\usepackage[title]{appendix}%
\usepackage{xcolor}%
\usepackage{textcomp}%
\usepackage{manyfoot}%
\usepackage{booktabs}%
\usepackage{algorithm}%
\usepackage{algorithmicx}%
\usepackage{algpseudocode}%
\usepackage{listings}%
\usepackage{natbib}
\usepackage{comment}
\usepackage{pdfpages}
\usepackage{hyperref}  

\usepackage{lipsum} 
\usepackage{url}  
\usepackage{doi}
\title{Graph Neural Network based Handwritten Trajectories Recognition}
\newif\ifuniqueAffiliation
\uniqueAffiliationtrue

\ifuniqueAffiliation 
\author{ 		  
	\href{https://orcid.org/0000-0000-0000-0000}{\includegraphics[scale=0.06]{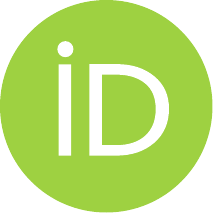}
	\hspace{1mm}Anuj Sharma}\\
	Department of Computer Science and Applications\\ 
	Panjab University, Chandigarh, India\\
	\texttt{anujs@pu.ac.in} \\
	\And
	\href{https://orcid.org/0000-0000-0000-0000}{\includegraphics[scale=0.06]{orcid.pdf}
	\hspace{1mm}Sukhdeep Singh} \\
	D.M. College (Affiliated to Panjab University, Chandigarh)\\
	Moga, Punjab, India\\
	\texttt{sukha13@ymail.com} \\
	\And
	\href{https://orcid.org/0000-0000-0000-0000}{\includegraphics[scale=0.06]{orcid.pdf}
	\hspace{1mm}S Ratna} \\
	Department of Computer Science and Applications\\
	Panjab University, Chandigarh, India\\	
}
\else
\usepackage{authblk}

\newbox{\orcid}\sbox{\orcid}{\includegraphics[scale=0.06]{orcid.pdf}} 
\author[1]{%
	\href{https://orcid.org/0000-0000-0000-0000}{\usebox{\orcid}\hspace{1mm}Anuj Sharma\thanks{\texttt{anujs@pu.ac.in}}}%
}
\author[2]{%
	\href{https://orcid.org/0000-0000-0000-0000}{\usebox{\orcid}\hspace{1mm}Sukhdeep Singh\thanks{\texttt{sukha13@ymail.com}}}%
}
\author[1]{%
	\href{https://orcid.org/0000-0000-0000-0000}{\usebox{\orcid}\hspace{1mm}S Ratna}
}

\affil[1]{Department of Computer Science and Applications, Panjab University, Chandigarh, India}
\affil[2]{D.M. College (Affiliated to Panjab University, Chandigarh),
	Moga, Punjab, India}
\fi

\renewcommand{\shorttitle}{\textit{arXiv} Template}

\hypersetup{
	pdftitle={A template for the arxiv style},
	pdfsubject={q-bio.NC, q-bio.QM},
	pdfauthor={David S.~Hippocampus, Elias D.~Striatum},
	pdfkeywords={First keyword, Second keyword, More},
}

\begin{document}

\maketitle

\begin{abstract}
The graph neural networks has been proved to be an efficient machine learning technique in real life applications. The handwritten recognition is one of the useful area in real life use where both offline and online handwriting recognition are required. The chain code as feature extraction technique has shown significant results in literature and we have been able to use chain codes with graph neural networks. To the best of our knowledge, this work presents first time a novel combination of handwritten trajectories features as chain codes and graph neural networks together. The handwritten trajectories for offline handwritten text has been evaluated using recovery of drawing order, whereas online handwritten trajectories are directly used with chain codes. Our results prove that present combination surpass previous results and minimize error rate in few epochs only.

\end{abstract}
\section{Introduction}
\label{sec1}
One of the Artificial Intelligence (AI) important applications is human handwritten text recognition. The Handwriting Recognition (HWR) refers to recognizing handwriting through machines. The handwritten text scanned and recognized is offline HWR in nature, whereas recognizing while writing is online HWR \cite{ijdar2017}. The handwriting trajectory refers to handwritten strokes which are set of sequential pixels in online HWR and set of pixels in offline HWR \cite{hwrreconstruct2019}. In offline HWR, these trajectories writing orders recovered through drawing order techniques, and in online HWR, the digital pen strokes refer to trajectories \cite{vjcs2015}. In either case, these trajectories are important sources of information to understand and recognize handwriting. Handwriting trajectories are also understood as the paths traced by a writing pen or stylus movements across a writing surface. These trajectories capture the spatial and temporal aspects of handwriting, including the sequence of strokes, their direction, and the relative positioning of characters \cite{trajectoryrecovery2023}. The trajectories in online HWR include important information as start or end of movements, velocity of pen movement, curvature, graphical shapes, time series information, dynamic nature of pen movements and behavioral part of writing. Today advances in digital pen technologies made it easier to capture and analyze handwriting trajectories for use in different domains. The chain code features are an important feature extraction technique related to trajectories and a popular way to recognize
these trajectories \cite{recoveryhwr2013}. These chain codes describe the boundary of the object by encoding the sequence of directions or transitions between neighboring points along the trajectory \cite{chaincode2019}. The chain code formed on the basis of boundary tracing using techniques as Huffman chain codes \cite{huffmancc2005}. The sequence of chain codes could be the original direction of trajectory or manually chosen points. Chain code features offer a compact representation of object boundaries and are invariant to translation, rotation, and scaling transformations.
\par The Graph Neural Networks (GNNs) are a class of neural networks designed to process and analyze data structured as graphs, also referred to as machine learning with graphs \cite{grlbook}. The nodes and edges are the main components of a graph. The chain code information for trajectories are transformed in the form of nodes and edges information. This way we can have offline and online HWR data in form of graphs and it could be applicable to GNN. Further, any data that could be well represented in the form of graphs for GNN may be used as an alternative to traditional machine learning approaches \cite{gnnsurvey2021}. The other important parts of GNN as message passing, graph convolution, aggregation function and pooling complete the recognition process of handwritten trajectories using GNN. The data could be graphically rich in raw form or could be converted to graphical form. The present work is related to HWR and raw form of handwritten strokes or trajectories are graphically rich in nature. Therefore, GNN could be a suitable choice in this field where input is in the form of chain code based trajectories.\\
Based on properties of HWR trajectories, chain codes and GNN, "Is it possible to implement a deep learning based approach using GNN to accept graphs derived from handwritten strokes or trajectories?". Our claim to this question is feasible based on experimental results performed on benchmarked datasets. The main contributions of this study are:
\begin{itemize}
	\item The present work could be first in this direction to use handwriting trajectories, chain codes and GNN together.
	\item The various GNN operators are explored to experiment in order to achieve state-of-the-art results.
	\item The approach is simple and Mathematically rich in nature. The hybrid nature of this approach strengthen the system working and indicate future use with other domain research problems.
	\item The benchmarked and inhouse datasets transformed to GNN requirements based on feature vectors.
\end{itemize}
Organization of rest of the paper: section \ref{2} include system overview and explanation of its components, section \ref{3} explain handwritten trajectories and chain code working to form feature vector. The section \ref{4} explain GNN working and Section \ref{5}  discuss experimentation. The last section \ref{6} conclude the paper.

\section{System Overview} \label{2}
This section explain overview of proposed system architecture, components, and their interactions. This explain system components functionality, and how different parts work together to achieve the desired outcomes. Our system include connection of different domains properties such as handwriting trajectories, graphs, chain codes and deep learning training. As happen in traditional AI approaches, we form dataset in desired form which split to train and test parts. The large dataset need special attention where batch loaders are used to train the data in batches. Our system accepts batch based data and reduce system complexities as happen in deep learning. The GNN dataset is special type of dataset in graphical form and any relational or hierarchical or other type of dataset need to be converted to graphical form of data \cite{gnnsurvey2021, gnnreviewAIopen2020, anujgnnreview2023}. Our system explained in this section collect graph properties and results in graphical dataset.
\par
\begin{figure*}[h]
	\begin{center}
		\includegraphics{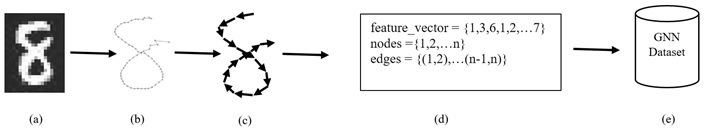}
		\caption{GNN based Dataset formation}
		\label{hwr2ds}
	\end{center}
\end{figure*}

\begin{figure*}[h]
	\begin{center}
		\includegraphics{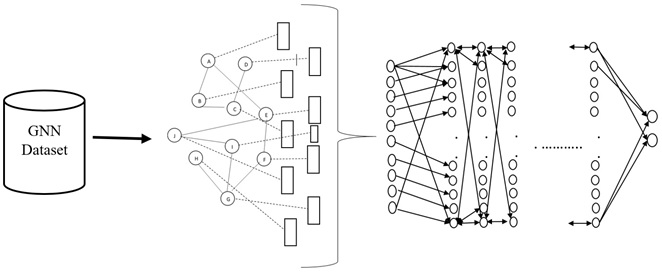}
		\caption{GNN training}
		\label{ds2gnn}
	\end{center}
\end{figure*}

 The proposed system design has been presented in figures \ref{hwr2ds} and \ref{ds2gnn}. The figure \ref{hwr2ds} include five stages from (a) to (e). The first stage (a) collect scanned image and transform to desired drawing order as shown in stage (b). The stage (b) directly implemented in case of online HWR strokes as drawing order is available during data capturing. The stage (c) is the chain code implementation and output in small segments, where each segment is a direction code. This is the simple graph nature where nodes are the segments start and end points and edges are links between these nodes. This way we have been able to collect information related to graphs for their respective nodes, edges and other features as degree of nodes, isolated nodes or self loops etc, as shown in \ref{hwr2ds} stage (d). Finally, GNN dataset formed as depicted in stage (e).
\par After GNN dataset stage, the figure \ref{ds2gnn} explain GNN working. The inside working of GNN is complex in nature. The GNN operators are the important part of GNN which include idea of convolution into graphs. This transformation is latent node features from node features. Further, various aggregation functions could be applied such as mean, sum or trainable layers \cite{graphkernelsurvey2021}. The input feature matrix at layer $l$ is $\mathbf{H}^{(l)} \in \mathbb{R}^{N \times d^{(l)}}$
with $N$ nodes and $d^{(l)}$ feature dimensions, the learnable weight matrix at layer $l$ is $\mathbf{W}^{(l)} \in \mathbb{R}^{d^{(l)} \times d^{(l+1)}}$, and the adjacency matrix of the graph is $\hat{\mathbf{A}} = \mathbf{A} + \mathbf{I}N$ with added self-loops. $\hat{\mathbf{D}}$ is the degree matrix of $\hat{\mathbf{A}}$, defined as $\hat{\mathbf{D}}{ii} = \sum_j \hat{\mathbf{A}}_{ij}$, and $\hat{\mathbf{D}}^{-\frac{1}{2}}$ is the normalized degree matrix. This results in output as a new feature matrix $\mathbf{H}^{(l+1)} \in \mathbb{R}^{N\times d^{(l+1)}}$ at layer $l+1$, where $d^{(l+1)}$ is the number of output features. The pseudocode explained in \ref{algorithm1} include working of figures \ref{hwr2ds} and \ref{ds2gnn} in sequence.

\begin{algorithm}
	\caption{Working of HWR chain codes using GNN (\textit{\textbf{Source code of Algorithm 1 given in Appendix I}})}  \label{algorithm1}	
	\begin{algorithmic}[1]		
		\State For offline HWR, collect images and transform to drawing order of images. In online HWR, the directly inputted handwritten strokes drawing order is available.
		\State The chain codes are derived for above step 1.
		\State The graph properties as nodes, edges and other dependent parameters are extracted. This information is now paart of GNN dataset.
		\State Import graph structured data and split as train and test data;
		\State Initialize model parameters as layers and its connections;
		\State Initialize model GNN using desired operator;
		\State Apply loss function and optimizer function to model GNN;
		\State Train the data using model GNN for number of epochs or threshold level;
		\State Predict the model for test data;
	\end{algorithmic}
\end{algorithm}

\section{Chain Code based Feature Extraction} \label{3}
The process of capturing and representing characteristics from patterns is referred to as feature extraction. These features help in analyzing the patterns for various applications such as handwriting recognition, image recognition, biometric verification etc. Common feature extraction techniques names are histogram of oriented gradients, scale invariant feature transform, local binary patterns, zoning and grid based features, statistical features, shape based structural features, chain code based features etc. The present study includes chain code feature extraction which has been used for offline HWR and online HWR both.
\par The offline HWR is scanned form of input handwritten text. The scanned form is an image and need preprocessing to extract features. The preprocessing include image conversion to binary images which are sent to thin image function to output in thin image \cite{binaryimagecc2002}. The thin image is a skeleton line form of original image. The thin image pixels list need attention to choose start point and sequential order of pixels, which is called recovery of drawing order. The recovery of drawing order technique has been explained in detail in literature \cite{recoveryhwr2013}. In this technique, start point is point with one neighbor and top left position in thin image. The next point is selected as next neighbor or the point among available points that closely fall in the previous direction. All the points in the pixel list are covered and drawing order of handwritten trajectory is recovered. The next step is formation of direction chain codes based on drawing order \cite{vjcs2015}. In online HWR, recovery of drawing is not required as input handwritten text is in form of sequential order of pixels. Thus, directly direction chain codes are formed \cite{ijdar2017}.
\par The direction chain codes include the segments based on pixels order. These segment directions are evaluated on the basis of directions. We have considered eight directions and results in chain code features for the input pixel list. The figure \ref{hwr2ds} stages (a) to (c) explain input image to chain code segments. These segments are assigned directions which results in chain code feature vector. In stage (c) of figure \ref{hwr2ds}, end points of each segments are nodes and link between these end points are edges. These nodes and edges are the two main parts of a graph as $G=(V,E)$, where $V$ is set of nodes and $E$ is the set of edges. The graph based on resultant feature vector is the input data for recognition stage using GNN.

\section{GNN} \label{4}
The graph representation learning is field of machine learning to work with graph structured data and GNN is the technique that work with graph structured data. Recently, the success of GNN has been witnessed in applications such as computer vision, recommender systems, drug discovery, social network and patterns classifications \cite{dlsurvey2022}. It is a type of neural network with deep learning features and graphical properties that include multiple layers, message passing mechanism and minimizing loss in each iteration. The propagation of information between neighboring nodes in multiple iterations is part of message passing  \cite{layer1000s2021} \cite{gnncomputation2008}. This was final layer output refers to various downstream tasks as node classification, link prediction or graph classification. The important intrinsic property of GNN is its rich Mathematical working which provide stepwise analysis of GNN working and graphically representation. The relationships of nodes, edges and graphs for their various properties outcome in domain level understanding. One important part of GNN is their operator function which is used to transform the features and structure of graph. This is possible with the message passing that exchange information between neighboring nodes \cite{node2022}. The operator commonly include convolution, pooling, attention, heterogeneous and point cloud functions \cite{pooling2022, aggregation2020}. The GNN ability to capture both local and global structural information that help to understand basic building components and overview level working simultaneously \cite{theoreticalgnn2022}. Some of the graphical rich data problems are possible to classify with GNN which were not possible using traditional machine learning methods.
\par The recent literature suggest that the field of GNNs is constantly evolving, with new architectures, algorithms, and its implementation in real life applications. The attention based mechanisms in GNN  which allow for more fine-grained control over the information flow between nodes and edges in the graph \cite{attention2022, anujijdar2023}. Also, the spatial-temporal modeling ability to model spatial and temporal information in graph-structured data. This way it can capture the dynamic nature of the underlying data and results in more accurate predictions and representations. Similarly, closely related meta-learning technique also train GNNs, allowing them to quickly adapt to new domains with labeled data. The GNN important feature as scalability which help to scale to large graphs in efficient way. Also, graph partitioning and sparsification reduce complexity of large graphs.
\par Therefore, in view to see above properties of GNN, it has been implemented in pattern recognition problem as handwriting recognition, where trajectories are considered as input data, which are graphically rich in nature. The handwritten trajectories information satisfy all major components of GNN working as nodes, edges, GNN operators, layers connections, pooling and aggregation functions. The GNN operators are of various nature and suitability to respective domain data is subject to experimentation. We explored various operators from literature and implemented for handwritten data. The common GNN operators working has been discussed in literature \cite{anujgnnreview2023}.

\section{Results} \label{5}
In this section, we have presented the experimental results for benchmark datasets. As our study include both online and offline HWR, we have used mnist dataset \cite{mnist} for offline HWR, where handwritten trajectories are recovered. For online HWR dataset, we have used unipen \cite{unipen} and indic handwritten strokes \cite{ijdar2017} datasets as mentioned in literature. As GNN is deep learning technique and need machines to computes lengthy calculations, we have used concept of batch loaders that train and test dataset in batches. This results in fast computational time and better handling of memory resources. As discussed in previous sections, GNN need dataset in graphical form, we have converted the datasets in same line of action. The formation of GNN dataset source code has been included in Supplementary file.
\par The offline HWR dataset as mnist has been used for experimentation with respect to recovery of drawing order technique. This dataset include 10 classes as digits from 0 to 9. We have used feature vector length as 41 and this feature vector include chain codes only. The details of these feature description has been discussed in literature \cite{vjcs2015}. The mnist include 60000 images of handwritten numerals. Therefore, we have formed 60000 respective graphs from the feature vectors. Similarly, 10000 test images refers to respective 10000 test graphs in mnist. Our architecture include three convolution steps and 16 hidden channels. The first step convolution includes feature vector to hidden channels, second step hidden channels to hidden channels and third step from hidden channels to number of classes. The optimizer has been used as adam with learning rate of 0.01. The mnist dataset has been used extensively in literature and error rates were reported as low as $0.19\%$ \cite{suen2012}. We notice that our work perform subject to performances of GNNs for mnist as error rate $0.86\%$ \cite{chebconv2016}.
\par One of the online HWR dataset as Gurmukhi HandWritten Test (GHWT) has been implemented as discussed in literature \cite{gurmukhidataset2017} \cite{ijdar2017}. It includes 62 classes and feature vector length is 25. We have used same number of hidden channels 16 as in offline case and three convolution steps with adam optimizer and learning rate as 0.01. The other online HWR dataset used as unipen \cite{unipen}. This dataset include online handwritten digits strokes for the ten classes from 0 to 9. Our results shows that GNN outperform literature results in both cases of offline and online. The train and test part of respective datasets are randomly shuffled for all classes to ensure unbiased outcomes. The table \ref{tab1} depict results for both offline and online HWR datasets, it clearly shows that our present approach outperform literature results for same feature vectors using GNN. Also, the number of epochs in GNN experimentation are moderate. We did GNN experimentation to the epochs number where we surpassed previous results in first few epochs. The GNN operators option is one advantage to get best results as not all operator perform same. Similarly, in our case, DeeperGCN GNN operator \cite{deepergcn2020} results for these chain code based feature vectors are same or better as compared to other GNN operators. The experiments of GNN operators with common parameters is other important factor has been considered.
\begin{table*}
	\begin{center}
		\caption{\textbf{Chain codes based handwitten trajectories recognition using GNN}} \label{tab1}
		\begin{tabular}{ |c|c|c|c| }
			\hline
			\textbf{Dataset} & \textbf{Classifier} & \textbf{Error rate (\%)}\\
			\hline
			mnist & GNN (this paper) & 1.10 (in 160 epochs)\\ \hline
			unipen & SVM \cite{ijdar2017} &  3.31 \\
			unipen & HMM \cite{ijdar2017} &  2.71 \\
			unipen & GNN (this paper) & 1.55 (in 110 epochs)\\ \hline
			GHWT & SVM \cite{ijdar2017} &  14.82 \\
			GHWT & HMM \cite{ijdar2017} &  13.20 \\
			GHWT & GNN (this paper) &  6.39 (in 100 epochs) \\
			\hline
		\end{tabular}
	\end{center}
\end{table*}

\par The GNN do have limitations as a classifier One of the challenge is over-smoothing that happen in deep architecture. Here, information is propagated through multiple layers node representations tend to become more similar in losing discriminative power. Also, GNNs are computationally expensive and require high end machines to handles big data. The computational cost is still high when we compare to other classifiers for medium size datasets but accuracy is the other useful outcome that prefers GNNs. One of the deep learning architecture is its black-box nature and GNN are again black-box appearance which lack interpretability compared to traditional graph algorithms. The other challenge is its perturbations in structures, a small change in graphs with respect to nodes or edges could result in variation of outcomes. The sparsity and irregularities further make task challenging for classification. In our case, handwritten trajectories especially through drawing order is first such work using GNN in this direction and these challenges could not dominate the entire classification progress.
This work has opened many future directions in connection to chain code and trajectory formation. There are many real life data challenges where chain code based graph understanding could be useful, especially large scale graphs complex nature. The rich Mathematical GNN working used to generate theoretically feasible and diverse solutions. Further, transfer learning with GNN could optimize existing solutions or feature extraction techniques. In view to see present work outcomes, GNNs shown to be effective for handwritten trajectories using traditional method chain code. The interesting part is the drawing order formation in offline handwriting.

\section{Conclusion} \label{6}
In this paper, novel offline and online HWR recognition technique has been proposed using chain code feature vectors and GNN, where each handwritten trajectory has been understood as a graph. The results are better as compared to literature where same dataset and feature vector used. We have received improvement in accuracy rates for all the datasets. Our results surpassed previous results in few epochs only and system is computationally efficient as batch loader based experimental setup used. The GNN requirement of dataset in graphical parameters form has given a new future direction for handwritten trajectories where graphs present simple understanding of data. The use of recovery of drawing order for offline handwritten text using GNN is first in this direction and results are encouraging. Further, online HWR results also support chain code and GNN combination. The work done in this study has opened new scope to extend present system and implementation in other domains too with similar feature sets.
\bibliographystyle{unsrtnat}
\bibliography{gnnhwr}

\end{document}